\documentclass[letterpaper, 10 pt, conference]{ieeeconf} 
\usepackage{times}
\usepackage{comment}
\usepackage{epsfig}
\usepackage{graphicx}
\usepackage{amsmath}
\usepackage{amssymb}
\usepackage{graphicx}
\usepackage{url}
\usepackage{array}
\usepackage{multirow}
\usepackage{subcaption}
\usepackage{color}
\usepackage{fancyhdr}
\usepackage{pifont}
\usepackage{multirow}
\usepackage{times}
\usepackage{amsmath}
\usepackage{amssymb}
\usepackage{array}
\usepackage{color}
\usepackage{mathtools}
\usepackage{comment}
\usepackage{url}
\usepackage[mathscr]{eucal}
\DeclareMathOperator*{\argmax}{argmax}
\usepackage[noend]{algpseudocode}
\usepackage{algorithm}
\usepackage{bm}
\usepackage{breqn}
\usepackage{color}

\fancypagestyle{specialfooter}{%
  \fancyhf{}
  
  \fancyfoot[L]{\textit{Please cite this article as: S.H.S. Basha, S.K. Vinakota, V. Pulabaigari et al.,  
AutoTune: Automatically Tuning Convolutional Neural Networks for Improved Transfer Learning, Neural Networks (2020), https://doi.org/10.1016/j.neunet.2020.10.009}}
}

\IEEEoverridecommandlockouts                              
\title{AutoTune: Automatically Tuning Convolutional Neural Networks for Improved Transfer Learning
}

\author{S.H. Shabbeer Basha, Sravan Kumar Vinakota,  Viswanath Pulabaigari, Snehasis Mukherjee, Shiv Ram Dubey \\
Indian Institute of Information Technology Sri City, Chittoor, India
}

\begin{document}
 \newcommand{\etal}{\textit{et al}. }
\newcommand{\cmark}{\ding{51}}%
\newcommand{\xmark}{\ding{56}}

\maketitle
\thispagestyle{specialfooter}
\pagestyle{empty}

\begin{abstract}
Transfer learning enables solving a specific task having limited data by using the pre-trained deep networks trained on large-scale datasets.
Typically, while transferring the learned knowledge from source task to the target task, the last few layers are fine-tuned (re-trained) over the target dataset. However, these layers are originally designed for the source task that might not be suitable for the target task. In this paper, we introduce a mechanism for automatically tuning the Convolutional Neural Networks (CNN) for improved transfer learning. The pre-trained CNN layers are tuned with the knowledge from target data using Bayesian Optimization. First, we train the final layer of the base CNN model by replacing the number of neurons in the softmax layer with the number of classes involved in the target task. Next, the CNN is tuned automatically by observing the classification performance on the validation data (greedy criteria). To evaluate the performance of the proposed method, experiments are conducted on three benchmark datasets, e.g., CalTech-101, CalTech-256, and Stanford Dogs. The classification results obtained through the proposed AutoTune method outperforms the standard baseline transfer learning methods over the three datasets by achieving $95.92\%$, $86.54\%$, and $84.67\%$ accuracy over CalTech-101, CalTech-256, and Stanford Dogs, respectively. The experimental results obtained in this study depict that tuning of the pre-trained CNN layers with the knowledge from the target dataset confesses better transfer learning ability. The source codes are available at \textcolor{blue}{\url{https://github.com/JekyllAndHyde8999/AutoTune_CNN_TransferLearning}}.
\end{abstract}

\section{Introduction}
\label{sec1}
The ability of Convolutional Neural Networks (CNN) for feature extraction and decision making in one-shot creates enormous demand in several application areas, such as
object recognition \cite{krizhevsky2012imagenet}, language translation \cite{zhang2015deep}, and many more. However, the performance of the deep learning models is sensitive w.r.t. the small changes made in both network hyperparameters settings, such as the number of layers, filter dimension of a convolution layer, etc. and it is also sensitive to the other training parameters, such as learning rate, activation function, and so on. Most of the CNNs available in the literature are carefully designed in terms of these hyperparameters by the domain experts \cite{krizhevsky2012imagenet, simonyan2014very, szegedy2015going, he2016deep}.

In recent years, researchers have made substantial efforts to automatically learning the structure of a CNN for a specific task \cite{zoph2018learning, liu2018progressive}, known as Neural Architecture Search (NAS). Although these methods out-perform most of the hand-engineered architectures, the search process requires huge computational resources and time to train the proxy CNNs that are explored during the architecture search process \cite{zoph2016neural}. Particularly while working on small datasets, the process of NAS becomes a challenge. Transfer learning is a popularly adopted technique to reduce the demand for both large computational resources and training data by providing promising performance over small datasets. 
\begin{figure*}[!t]
    \centering
    \includegraphics[width=0.8\textwidth]{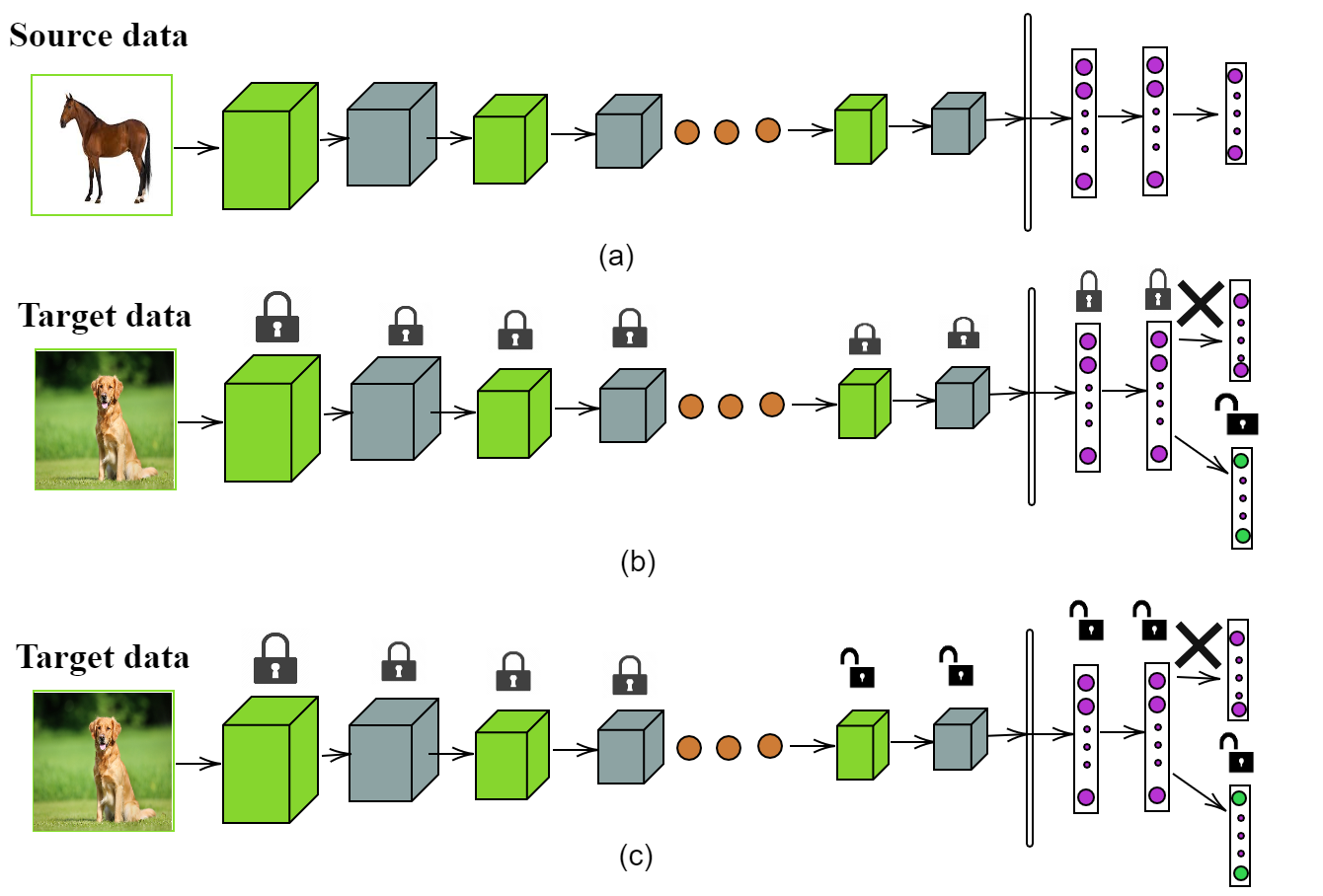}
    \caption{Overview of the proposed method. a) Typically, the CNN models used for transferring the knowledge are initially trained over a large-scale image dataset. b) Conventionally, while transferring the learned knowledge from the source task to the target task, the last one or a few layers of the pre-trained CNN are fine-tuned over the target dataset. c) The proposed AutoTune method tunes the $k$ number of layers automatically using Bayesian Optimization \cite{frazier2018tutorial}. Note that the lock and unlock symbols are used to represent the frozen and fine-tuned layers, respectively. Finally, the tuned CNN layers can be re-trained over the target dataset for improved transfer learning. Different colors represent different CNN layers.}
    \label{fig:motivation}
\end{figure*}

Many machine learning algorithms assume that the training (source) data and future data (target) have the same distribution. In this direction, many Metric-learning algorithms \cite{hu2015deep,dong2017dimensionality} are proposed in the literature. To mention a few, \textit{Deep Transfer Metric Learning (DTML)} method introduced by Hu \etal \cite{hu2015deep}, in which a discriminative distance network is trained for cross-dataset visual recognition by maximizing the inter-class variations and minimizing the intra-class similarity. Similarly, Dong \etal \cite{dong2017dimensionality} proposed an \textit{Ensemble Discriminative Local Metric
Learning (EDLML)} which aims at learning a sub-space to keep all the intra-class samples as close as possible, while the samples belong to different classes are well separated. Shi \etal \cite{shi2015domain} presented a semi-supervised domain adaptation approach which finds new representations of the images belong to the source domain using multiple linear transformations.

Typically, while transforming the learned knowledge from source task to the target task, the classification layer of the pre-trained CNN is dropped, after which a new softmax layer is stacked that is trained over the target dataset during transfer learning. The number of layers to be fine-tuned can be decided majorly based on the size of the target dataset and the similarity between the source and target datasets \cite{karpathy2017cs231n}. However, the pre-trained CNN model is designed for the source dataset, which may not perform well over the target dataset. 

In this paper, we attempt to automatically tune the CNN to make it suitable for the target task/dataset. To achieve this objective, initially, we drop the softmax layer of the pre-trained CNN by replacing it with a new softmax layer having the neurons equal to the number of classes in the target dataset. 
Next, we automatically tune the layers of CNN using Bayesian Optimization \cite{frazier2018tutorial}.
It is a well-known fact that the initial layers of CNN represent primitive features such as edges and blobs which are generic to many tasks. On the other hand, the final layers of CNN represent the features that are very specific to the learning task \cite{zeiler2014visualizing}. Based on the above idea, in our work, the layers of the CNN are tuned from right to left (i.e., from final layers to initial layers) by observing the network performance on the validation data. The results obtained through our experiments indicate that tuning the optimum number of layers with respect to the target task leads to better image classification performance in the context of transfer learning. Fig.\ref{fig:motivation} shows an overview of the proposed idea of improving the transfer learning process. Next, we provide a survey of literature in the specific research area focused in this study.

\section{Related Works}
The hierarchical feature extraction ability of deep neural networks enables the adoption of a deep network. The pre-trained network over a large-scale dataset (source task) can be utilized to solve the specific task/ problem (target task) through transferring the knowledge learned from the source task. Transfer learning has been widely adopted in domains where collecting the annotated examples is expensive (labor-intensive) and time-consuming task, such as biomedical \cite{raghu2020eeg,shin2016deep}, agriculture \cite{kamilaris2018deep}, signature generation \cite{nahmias2020deep} and many more. Khan \etal \cite{khan2019novel} proposed an average-pooling based classifier to detect and classify breast cancer images. Han \etal \cite{han2018new} introduced a two-step method to improve the generalization ability of CNN over the target dataset using web data augmentation. Yosinski \etal \cite{yosinski2014transferable} conducted a study to measure the generality and specificity of different neurons involved in a deep neural network. In other words, they introduced a mechanism to quantify the transfer-ability of features learned by each layer of a CNN.  Wang \etal \cite{wang2019domain} introduced an approach to induce a common representation feature space for both source and target domains using a CNN model. A semi-supervised domain adaptation approach proposed by Shi \etal \cite{shi2015domain}, which finds new representations of the images belong to different classes from the source domain.


Automatically searching for better performing CNN architectures for a given task (also called NAS) has gained interest among the researchers in recent years \cite{elsken2019neural}. Before the emergence of NAS, hyperparameter optimization has achieved great success in tuning the machine learning algorithms \cite{bergstra2011algorithms, snoek2012practical}. NAS-based CNNs available in the literature consume an enormous amount of search time and GPU hours to find better performing CNNs. For instance, Reinforcement Learning (RL) based NAS method proposed by Zoph \etal \cite{zoph2016neural} trained $12,800$ proxy CNN models for 28 days using 800 GPUs. Moreover, NASNet \cite{zoph2018learning} utilized $500$ GPUs for $4$ days to train $20,000$ proxy CNN models that are sampled during the architecture search. Very recently, Jiang \etal \cite{jiang2020efficient} proposed a multi-objective NAS that is intended to minimize both classification error and network parameters.

Transfer learning allows the applications to utilize the advantage of deep neural networks by reusing the learned knowledge from the source task. Fine-tuning the pre-trained CNN over the target dataset may produce satisfactory results. However, the pre-trained CNN is designed for the source task which may not be suitable for the target task, especially the deeper layers. Therefore, tuning the optimal number of CNN layers and suitable hyperparameters with the knowledge from the target dataset may produce better results compared to the traditional fine-tuning approaches over the target dataset. 

After transferring the learned knowledge from a source task, the capacity of the network increases for the target task \cite{molchanov2016pruning}, if the target dataset has a very less number of training examples. Molchanov \etal \cite{molchanov2016pruning} proposed a mechanism to progressively prune the least important feature maps to reduce the model's capacity while fine-tuning over the target task. Various studies have reported that the CNN representations learned from a large-scale image dataset in a source domain, can be successfully transferred to a target domain \cite{yosinski2014transferable, azizpour2015factors}. However, the target domain has limited data compared to the source domain in practice. In such scenarios, transfer learning suffers from overfitting. To address this problem, Liu \etal \cite{liu2017sparse} introduced a framework called Hybrid-TransferNet to improve the network's generalization ability while transferring the knowledge from the source domain to the target domain by removing the redundant features. Similarly, Ayinde \etal \cite{ayinde2019redundant} proposed a method to reduce the inference time of deep neural networks by pruning the redundant feature maps (filters) based on the relative cosine similarities in the feature space.

Recently, Basha \etal \cite{basha2020autofcl} proposed a framework called AutoFCL for automatically tuning the Fully Connected (FC) layers of a pre-trained CNN with the knowledge from the target dataset while transferring the knowledge from the source task. However, in this work, the search space is limited to fully connected layers. To extend this work, we design a mechanism for automatically tuning the CNN w.r.t. the target dataset for improved transfer learning.  


The contributions of this research can be summarized as follows:
\begin{itemize}
 \item This paper introduces a framework called AutoTune which finds the number of layers to be fine-tuned automatically for a target dataset for improved transfer learning.
 \item Bayesian Optimization technique is applied to learn the pre-trained CNN layers with the knowledge of the target dataset.
 \item Several experiments are conducted over CalTech-101, CalTch-256, and Stanford Dogs datasets to justify the efficacy of the proposed model using the popular pre-trained CNNs, namely, VGG-16 \cite{simonyan2014very}, ResNet-50 \cite{he2016deep}, and DenseNet-121 \cite{huang2017densely}. The obtained results are compared with state-of-the-art that includes both transfer learning and non-transfer learning-based methods.
\end{itemize}

Next, we discuss the proposed method in detail.

\section{Methods}

\begin{algorithm*}
\caption{AutoTUNE: Automatically tuning the CNN for improved transfer learning using Bayesian optimization} 
\textbf{Inputs:} $B$ (Base$_{CNN}$),  $HParam\_space$ (hyperparameters search space), Training$_{Data}$, Validation$_{Data}$, Epochs (num of epochs to train each proxy CNN). \\
\textbf{Output:} A CNN with target-dependent network architecture.

\label{Auto_Tune_algorithm}
\begin{algorithmic}[1]
 \Procedure{AutoTune}{} 
 \State Assume Gaussian Process (GP) prior on the objective function $F$
 \State Find and observe the objective $F$ at initial $m_0$ points
 
 
 
 
 
$n = m_{0}$ 
 
\While{$k \in n+1, .., N$}   \Comment{explore the hyperparameter search space}

\State Update the posterior distribution on $F$ using the prior
\State Choose next sample $x_k$ that maximizes the acquisition function value

 
\State Observe $y_k = F(x_k)$
\EndWhile
  
\State \textbf{return} $x_k$\Comment{return a point with best FC layer structure}

\EndProcedure

\end{algorithmic}
\end{algorithm*}

The task of automatically tuning a pre-trained CNN w.r.t. the target dataset can be formulated as a black-box optimization problem where we do not have direct access to the objective function. In this paper, tuning the CNN layers with the knowledge of the target dataset for improved transfer learning is achieved using Bayesian Optimization \cite{frazier2018tutorial}. Let $F$ be the objective function, which is of the form,
\begin{equation}
    F: \mathbb{R}^d \xrightarrow{} \mathbb{R}.
\end{equation}

The objective of the Bayesian optimization can be represented mathematically as follows, 
\begin{equation}
    x_* = \argmax_{x \in \mathcal{S}} F(x),
\end{equation}
where $x\in\mathbb{R}^d$ is the input, the hyperparameter search space is denoted by $\mathcal{S}$ which is shown in Table \ref{hyperparameter_space_table}. Evaluating the value of the objective function $F$ at any point in the search space is expensive. It can be obtained by fine-tuning (re-training) the proxy CNN layers (explored during the architecture search) over the target dataset. 
\begin{table*}[!t]
\caption{The search space considered for the hyperparameters involved in the convolutional neural network layers such as Convolution, max/average-pooling, and dense layers.}
\centering
\begin{tabular}{|l|l|l|l|}
\hline
S.No. & Type of the layer                                                  & Parameters involved                                                                          & Parameter values                                                                                 \\ \hline
1    & Convolution                                                        & \begin{tabular}[c]{@{}l@{}}Receptive filter size\\ Stride\\ \#receptive filters\end{tabular} & \begin{tabular}[c]{@{}l@{}}\{1, 2, 3, 5\}\\ Always 1\\ \{64, 128, 256, 512\} \end{tabular}          \\ \hline
2    & Max-pooling                                                        & \begin{tabular}[c]{@{}l@{}}Receptive filter size\\ Stride\end{tabular}                       & \begin{tabular}[c]{@{}l@{}}\{2, 3\}\\ Always 1\end{tabular}                                     \\ \hline
3    & Average-pooling                                                    & \begin{tabular}[c]{@{}l@{}}Receptive filter size\\ Stride\end{tabular}                       & \begin{tabular}[c]{@{}l@{}}\{2, 3\}\\ Always 1\end{tabular}                                   \\ \hline
3    & \begin{tabular}[c]{@{}l@{}}Fully Connected or\\ Dense\end{tabular} & \begin{tabular}[c]{@{}l@{}}\# FC layers\\ \# neurons \\ \end{tabular}                          & \begin{tabular}[c]{@{}l@{}} \{1, 2, 3\} \\ \{64, 128, 256, 512, 1024\}\\ \end{tabular} \\ \hline
4    & \begin{tabular}[c]{@{}l@{}}Dropout\end{tabular} & \begin{tabular}[c]{@{}l@{}} dropout factor\end{tabular}                          & \begin{tabular}[c]{@{}l@{}} \ {[}0, 1{]} with offset 0.1\end{tabular} \\

\hline
\end{tabular}
\label{hyperparameter_space_table}
\end{table*}

The optimal configuration of the hyperparameters involved in the tuned layers (learned using Bayesian optimization) is represented using $x_*$. Hence, the pre-trained CNN having the last few layers with the optimal structure ($x_*$) is responsible for obtaining the best performance on the validation (held-out) data Validation$_{Data}$. The proposed method for automatically tuning the CNN with the knowledge from the target dataset yields improved performance while transferring the learned information from the source dataset to the target dataset. The proposed idea is outlined in Algorithm \ref{Auto_Tune_algorithm}.

\subsection{Proposed AutoTune Method}
The objective of the proposed AutoTune method is to find the optimal structure of pre-trained CNN for improved transfer learning. To achieve this objective, Algorithm \ref{Auto_Tune_algorithm} takes  Base$_{CNN}$ ($B$) i.e., a pre-trained CNN, hyperparameter search space (HParam$_{space}$), Training$_{Data}$, Validation$_{Data}$, and maximum number of Epochs ($E$) to train each proxy CNN as input. It tunes the pre-trained CNN layers w.r.t. the target dataset using Bayesian optimization (Bayes Opt) \cite{frazier2018tutorial}. 

Bayesian optimization has been used widely to tune the hyperparameters involved in machine learning algorithms such as deep neural networks \cite{snoek2012practical}. Bayesian optimization is a type of machine learning based optimization problem in which the objective is a black-box function. Here, we provide brief details about Bayesian optimization, for complete details we recommend the reader to refer the original paper \cite{frazier2018tutorial}. Bayesian optimization includes two key components, including a surrogate model and an acquisition function. The surrogate model is a Bayesian statistical model that builds an approximation for the objective function using Gaussian Process (GP) Regression \cite{rasmussen2003gaussian}. The acquisition function utilizes this Bayesian statistical model to make the search process productive by proposing the next point to explore from the search space. 

As outlined in Algorithm \ref{Auto_Tune_algorithm}, initially the objective function $F$ is observed at $m_0$ points which are chosen uniformly random (in our experimental settings $m_0$ is considered as $20$). Among the $m_0$ function evaluations, we pick the optimal point $x^+$, the hyperparameter configuration corresponding to the maximum validation accuracy ($F(x^+)$) observed while fine-tuning the CNN over the target dataset. Next, we need to conduct an evaluation at new point $x_{new}$, which is $F(x_{new})$. At this moment, the optimal objective value is either $F(x^+)$ if $F(x^+) \ge F(x_{new})$ or $F(x_{new})$ if $F(x^+) \le  F(x_{new})$. The improvement or gain in the value of objective after observing at this point is $F(x_{new})-F(x^{+})$ which is either positive when $F(x_{new}) > F(x^{+})$, or zero when $F(x_{new}) \le F(x^{+})$. We can represent the value of improvement as $[F(x_{new}) - F(x^+)]^+$, where $[]^+$ denotes the positive quantity of improvement. At iteration $k = m_0+1$, we choose $x_{new}$ at which the improvement is high. However, the value of objective $F(x_{new})$ is unknown before evaluation (which is a typically expensive task). Alternatively, we can compute the expected value of the improvement and choose $x_{new}$ for which the improvement is maximized. The Expected Improvement (EI) is defined as follows:
\begin{equation}
    EI_{m_{0}}(x) := E_{m_{0}} [[F(x_{new})-F(x^+)]^+]
\end{equation}
where $E_{m_{0}}$ is the expectation computed under posterior distribution given the evaluations of objective $F$ at the points $x_1$, $x_2$, ..., $x_{m_{0}}$. The posterior probability of $F(x_{new})$ given $F(x_{1:m_{0}})$ is normally distributed with mean $\mu_{m_{0}}(x_{new})$ and variance $\sigma^2_{m_{0}}({x_{new}})$, which can be formally represented as,
\begin{equation}
    F(x_{new})|F(x_{1:m_{0}}) \sim Normal(\mu_{m_0}(x_{new}),\sigma^2_{m_{0}}(x_{new}))
    \label{conditional_bayes}
\end{equation}
where $\mu_{m_{0}}(x_{new})$ and $\sigma^2_{m_{0}}(x_{new})$ are analytically computed as,
\begin{dmath}
\mu_{m_{0}}(x_{new}) = \sum_0(x_{new},x_{1:m_{0}})\sum_0(x_{1:m_{0}},x_{1:m_{0}})^{-1}(F(x_{1:m_{0}})
- \mu_{0}(x_{1:m_{0}})) + \mu_0({x_{new}})
\end{dmath}
\begin{dmath}
\sigma^{2}_{m_{0}}(x_{new}) = \sum_0(x_{new},x_{new})-\sum_0(x_{new}, x_{1:m_{0}})\sum_0(x_{1:m_{0}},x_{1:m_{0}})^{-1}\sum_0(x_{1:m_{0}},x_{new})
\end{dmath}

The Expected Improvement (EI) acquisition function lets the Algorithm \ref{Auto_Tune_algorithm} to evaluate at the point which results in a maximum improvement in the expectation. 
\begin{equation}
    x_{k+1} = \argmax EI_{k}(x)
\end{equation}
We update the posterior distribution upon observing the objective $F$ at each point using equation (\ref{conditional_bayes}). Throughout our experiments, the evaluations (N) of the objective F are performed for 50 (including 20 initial evaluations which are sampled in a uniform random manner) in the case of Bayesian Optimization. On the other hand, 100 evaluations are performed in case of random search.

\section{Architecture Tuning Search Space}
\label{hyperparameter_space}
Here, we provide a detailed discussion on the used search space for the hyperparameters involved in the different layers of a deep neural network, such as convolution, max-pooling, average pooling, and dense layers. In our experimental settings, we consider the plain as well as skip connection based CNNs. More concretely, we consider the popular CNNs such as VGG-16 \cite{simonyan2014very}, ResNet-50 \cite{he2016deep} and DenseNet-121 \cite{huang2017densely}. These networks are originally pre-trained over ImageNet \cite{deng2009imagenet} dataset  for automatically tuning w.r.t. the target dataset for improved transfer learning.

Basha \etal \cite{basha2020autofcl} observed that learning the structure of Fully Connected (FC) layers with the knowledge from the target dataset and fine-tuning these learned FC layers over the target dataset leads to better performance compared to just fine-tuning with the original CNN. To extend this work, in this article, we propose a framework for automatically tuning the CNN (beyond FC layers) for improved transfer learning. From the literature, we found that a shallow/deeper CNN, which has 3 FC layers  \cite{krizhevsky2012imagenet, simonyan2014very} including the output layer achieves comparable performance. Hence, we consider, the search space for the "number of FC layers" hyperparameter in the range \{1, 2, 3\} including the output FC layer. Another key hyperparameter involved in FC layers is the number of neurons, for which our search space is \{64, 128, 256, 512, 1024\}. Dropout \cite{srivastava2014dropout} is a popular regularization method adopted to generalize the network performance over unseen data. We employed dropout after each dense layer of proxy CNNs explored during the architecture search. The dropout rate is tuned within the range $[0, 1]$ with an offset $0.1$, i.e., the optimal dropout factor is learned from the values $\{0.0, 0.1, ... , 0.9, 1.0\}$. The hyperparameters involved in other layers such as Convolution, Pooling are shown in Table \ref{hyperparameter_space_table}. Our search space involves $6$ operations in both convolution and pooling layers which is less as compared to $8$ operations used in \cite{liu2018progressive} and $13$ operations used in \cite{zoph2018learning}.  

Throughout our experiments, we have not modified the connectivity of the CNN models that are utilized to carry out the experiments. We utilize two more operators, including 1) $1\times1$ convolution and 2) upsample operations (similar to un-pooling) to tackle the tensor dimension mismatch in depth and spatial dimensions, respectively.


\section{Experimental Settings}
We first demonstrate the details about the hyperparameters such as learning rate, optimizer, etc., employed while training the CNNs in Subsection \ref{training_details}. The pre-trained deep neural networks used to tune the CNN w.r.t. the target dataset are discussed in Subsection \ref{CNN_models}, and the datasets utilized for conducting the experiments are discussed in Subsection \ref{datasets}. 

\subsection{Training Details}
\label{training_details}
We call the CNNs explored during the search process as proxy CNNs. To train the proxy CNNs, the parameters or weights involved in the layers (which are tuned during the search process) are initialized with "He-normal" initialization \cite{he2015delving}. The CNNs are trained using Adagrad optimizer \cite{duchi2011adaptive} with an initial value of the learning rate as $0.01$. The learning rate is decreased by a factor of $\sqrt{0.1}$ if there is no reduction in the validation loss. Rectified Linear Unit (ReLU) is employed as the activation function throughout the experiments. Since the architecture search is a time-consuming task, to alleviate this situation each proxy CNN is trained for $50$ epochs. To reduce the over-fitting, data augmentation such as shearing, zooming the images, horizontal, and vertical flip image transformations are employed. Batch Normalization \cite{ioffe2015batch} is used after every dense layer to accelerate the training process and to increase the generalization ability.

\begin{table*}[!t]
\centering
\caption{Comparing the classification performance obtained using the proposed method which is developed using both Bayesian Optimization (BO) and Random Search (RS)  with state-of-the-art methods over CalTech-101, CalTech-256, and Stanford Dogs datasets. We compare our results with both transfer learning and non-transfer learning based methods. The best and second-best classification accuracies on each dataset are highlighted in \textbf{\textit{bold-italic}} and \textbf{bold}, respectively. Note that the results of existing methods are taken from the respective papers.}
 \resizebox{0.7\textwidth}{!}{%
\begin{tabular}{|c|l|l|l|}
\hline
\multicolumn{1}{|l|}{Dataset}        & Method                                                                                    & Accuracy       & Transfer Learning                              \\ \hline
\multirow{4}{*}{CalTech-101}        &  Lee \etal \cite{lee2009convolutional}                     & $65.4$    & \xmark                                                \\ \cline{2-4} 
                                     & Zeiler \etal \cite{zeiler2014visualizing}  & 86.5    & \cmark                                                    \\ \cline{2-4} 
                                     & Cubuk \etal \cite{cubuk2019autoaugment}                   & 86.9      & \cmark                                                    \\ \cline{2-4} 
                                     &  Sawada \etal \cite{sawada2019improvement}                  & 91.8      & \cmark                                                    \\ \cline{2-4} 
                                                  & Basha \etal  \cite{basha2020autofcl}                                                            &   94.38 $\pm$ 0.015       &  \cmark           \\ \cline{2-4} 
                                     & \textbf{AutoTune (BO) + VGG-16 }                                                       & \textbf{95.83 $\pm$ 0.004}           & \cmark           \\ \cline{2-4}
                                     &  AutoTune (RS) + VGG-16                                                       & 93.54 $\pm$ 0.02           & \cmark          \\ \cline{2-4}
                                     & AutoTune (BO) + ResNet-50                                                               & 93.57 $\pm$ 0.034           & \cmark            \\ \cline{2-4} 
                                   &  AutoTune (RS) + ResNet-50                                                               &  91.52 $\pm$ 0.004           & 
                                  \cmark            \\ \cline{2-4}             &
                        \textbf{\textit{ AutoTune (BO) + DenseNet-121}}                                                               & \textbf{\textit{ 95.92 $\pm$ 0.025}}        & \cmark            \\ \cline{2-4}                      
                        &       AutoTune (RS) + DenseNet-121                                                               & 93.71 $\pm$ 0.031          & \cmark         \\ \cline{2-4} 
                                    \hline
\multirow{6}{*}{CalTech-256} &  
Zeiler \etal  \cite{zeiler2014visualizing}                        & 74.2     & \cmark                                                \\ \cline{2-4} 
&
Wang \etal \cite{wang2016cost}                        & 74.2     & \xmark                                                \\ \cline{2-4} 
                                     &  Schwartz \etal \cite{schwartz2018delta}                              & $83.6$ & \xmark                                               \\ \cline{2-4} 
                                     & Chu \etal \cite{chu2016best}                     & $71.4$  & \cmark                                               \\ \cline{2-4} 
                                     &  Cai \etal \cite{cai2016probabilistic}                           & 83.3   & \cmark                                                   \\ \cline{2-4} 
                                     &  Zheng \etal \cite{zheng2016good}                & $83.27$    & \cmark                                                   \\ \cline{2-4} 
                                     & Mahmood \etal \cite{mahmood2017resfeats}                                                                   &    82.1       & \cmark               \\ \cline{2-4} 
                                     & 
                                     Wang \etal \cite{wang2019sharpen}                                                                & 81.32          & \cmark            \\ \cline{2-4} 
                                    & AutoTune (BO) + VGG-16                                                       & $82.47 \pm 0.003$           & \cmark           \\ \cline{2-4}
                                    & AutoTune (RS) + VGG-16                                                      & $81.83 \pm 0.021$          & \cmark           \\ \cline{2-4}
                                     & AutoTune (BO) + ResNet-50                                                               & 84.31 $\pm$ 0.005            & \cmark             \\ \cline{2-4} 
                                                               & AutoTune (RS) + ResNet-50                                                       & $83.74 \pm 0.001$           & \cmark           \\ \cline{2-4}           
                                     
                                     &
                                     \textbf{\textit{AutoTune (BO) + DenseNet-121}}                                                               & \textbf{\textit{86.54 $\pm$ 0.023}}           & \cmark          \\ \cline{2-4}  
                                     & \textbf{AutoTune (RS) + DenseNet-121}                                                      & \textbf{85.96 $\pm$ 0.045}           & \cmark           \\ \cline{2-4}            
                                    \hline

    \multirow{8}{*}{Stanford Dogs}          
    &

Murabito \etal \cite{murabito2018top}               & $70.5$ & \xmark                                             \\ \cline{2-4}
     &

Lee \etal \cite{lee2019learning}               & $55.2$ & \cmark                                              \\ \cline{2-4} 
    &

Li \etal \cite{li2020baseline}               & 77.1 & \cmark                                                \\ \cline{2-4} 
                                     &
                                  \textbf{Dubey \etal \cite{dubey2018pairwise}}                  & \textbf{83.75}    & \textbf{\cmark}                                      \\
\cline{2-4} 
&
Shen \etal \cite{shen2019amalgamating}        & 65.5  & \cmark                                                    \\ \cline{2-4} 
                                    & AutoTune (BO) + VGG-16                                                       & $81.23 \pm 0.002$          & \cmark           \\ \cline{2-4} 
                                     &
                         AutoTune (RS) + VGG-16                                                      & $78.21 \pm 0.005$          & \cmark           \\ \cline{2-4} 
                                     &           AutoTune  (BO) + ResNet-50                                                               & 83.17 $\pm$ 0.022           & \cmark            \\ \cline{2-4} 
                                     &
                         AutoTune (RS) + ResNet-50                                                               & 82.4 $\pm$ 0.06           & \cmark            \\ \cline{2-4} 
                                     &                          \textbf{\textit{AutoTune (BO) + DenseNet-121}}                              &  \textbf{\textit{84.67 $\pm$ 0.014}}          & \textbf{\textit{\cmark}}           \\
             \cline{2-4}                                 &
                         AutoTune (RS) + DenseNet-121                                                               & 83.19 $\pm$ 0.004           & \cmark            \\ \cline{2-4}
                                     \hline 
\end{tabular}
}
\label{results_comparison_stateoftheart}
\end{table*}

\subsection{Deep CNNs used for AutoTune}
\label{CNN_models}
From the literature of deep neural networks, we can broadly classify the CNNs into two types based on the network connectivity, i.e., i) Chain-structured or Plain CNNs \cite{elsken2019neural} and ii) CNNs with skip connections. 

\begin{figure}[!t]
    \centering
    \includegraphics[width=0.5\textwidth]{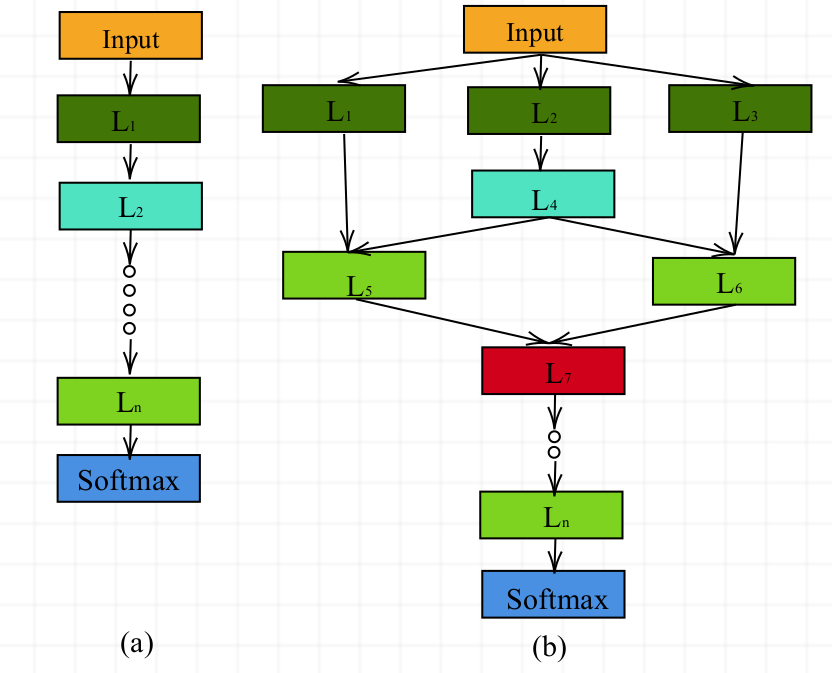}
    \caption{The pictorial representation of different types of CNNs. Each rectangle box represents a layer in CNN such as convolution, max-pooling, and dense layers. Different colors are used to represent different layers. a) Illustration of chain-structured CNNs. b) The more complex type of architecture involving skip connections, where a layer receives input feature-map from one or more layers.}
    \label{CNN_types}
\end{figure}

\textbf{Plain CNNs:} A plain CNN consists of collection of $n$ layers stacked sequentially. The $k^{th}$ layer $L_k$ receives the input from layer $L_{k-1}$ and its output feature map is supplied as input to the layer $L_{k+1}$. For instance, the hand-designed CNNs proposed in the initial years (2012-2014) such as LeNet \cite{lecun1998gradient}, AlexNet \cite{krizhevsky2012imagenet}, ZFNet \cite{zeiler2014visualizing}, VGGNet \cite{simonyan2014very}, and NAS based models \cite{baker2016designing, baker2017accelerating} have chain-structured connections. A typical structure of plain CNNs is shown in Fig. \ref{CNN_types}(a). In this article, we utilize the VGG-16 \cite{simonyan2014very}, a plain structured deep CNN for conducting the experiments. 

\textbf{CNNs with skip connection:} This class of CNNs involves skip-connections which allow a layer to receive input feature maps from more than one layer. 
Let us consider, a CNN is having $n$ layers which are represented as \{$L_1, L_2, ... , L_k, ..., L_{n-1}, L_{n}$\}, where the layers $L_1$, $L_n$ are input, output layers, respectively. In the CNNs involving skip connections, the layer $L_k$ receives input feature-maps from both layers $L_k-1$ and $L_k-2$ as in ResNet \cite{he2016deep}. Similarly, in DenseNet proposed by Huang \etal \cite{huang2017densely}, layer $L_k$ receives the input feature-maps from all of its previous layers $L_1, L_2, ....L_{k-1}$. In 2015, Szegedy \etal \cite{szegedy2015going} developed a CNN called GoogLeNet by investing more time and human efforts. Recently proposed NAS based CNNs such as \cite{liu2015content,zoph2018learning} discovers more complex CNNs during their search process. Fig \ref{CNN_types}(b) demonstrates the abstract view of CNNs involving skip connections. In this research, we use ResNet-50 \cite{he2016deep} and DenseNet-121 \cite{huang2017densely} for conducting the experiments of automatically tuning CNNs for improved transfer learning. 

\subsection{Datasets}
\label{datasets}
We use three publicly available standard datasets, including CalTech-101 \cite{fei2004learning}, CalTech-256 \cite{griffin2007caltech}, and Stanford Dogs \cite{khosla2011novel}, to perform the experiments.

\begin{figure*}[!t]
    \begin{center}
    \includegraphics[width=\textwidth]{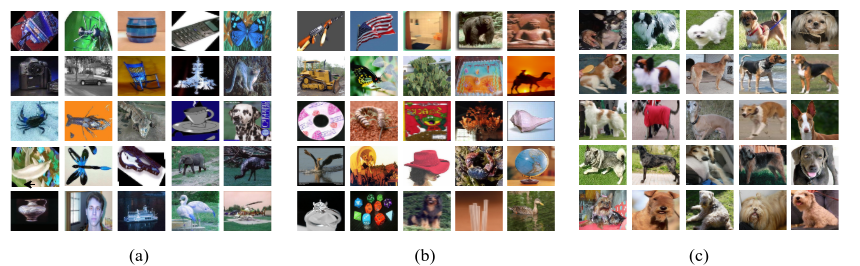}
    \caption{A set of random examples from a) CalTech-101 \cite{fei2004learning}, b) CalTech-256 \cite{griffin2007caltech}, and c) Stanford Dogs datasets \cite{khosla2011novel}. }
    \label{fig:dataset_samples}
    \end{center}
\end{figure*}

\subsubsection{CalTech-101}
The CalTech-101 dataset is proposed by Fei \etal \cite{fei2004learning}. It contains images from $101$ classes. This dataset has $9144$ images, among which $7315$ images are utilized for training and the remaining $1839$ images are used for validating the classification performance of the CNN models. The dimension of the images is adjusted to $224\times224\times3$ to fit the images as per the input needed to the deep neural networks. A few samples of images are depicted in Fig. \ref{fig:dataset_samples}(a). 

\begin{table*}[!t]
\centering
\caption{The best configuration of CNN layers learned for better transfer learning using Bayesian optimization. The base CNN model's layers are tuned w.r.t. the target datasets such as CalTech-101, CalTech-256, and Stanford Dogs. Note that the listed Fully connected (FC) layer's configuration does not include the output FC layer as it always has the number of neurons same as the number of classes.}
\label{tab:results_bayes_comparison}
\resizebox{\textwidth}{!}{%
\begin{tabular}{|c|c|c|c|c|c|l|l|c|l|c|}
\hline
\multirow{2}{*}{CNN}          & \multirow{2}{*}{Dataset}         & \multicolumn{3}{c|}{FC layer}                                                                            & \multicolumn{3}{c|}{Convolution layer}                                                    & \multicolumn{2}{c|}{Max-pooling layer}                                        & \multirow{2}{*}{\begin{tabular}[c]{@{}c@{}}Validation\\ Accuracy\end{tabular}} \\ \cline{3-10}
                              &                                  & \#layers              & \#neurons             & \begin{tabular}[c]{@{}c@{}}dropout\\ factor\end{tabular} & \#layers              & \begin{tabular}[c]{@{}l@{}}filter\\ size\end{tabular} & \#filters & \#layers              & \begin{tabular}[c]{@{}l@{}}filter\\ size\end{tabular} &                                                                                \\ \hline
\multirow{3}{*}{VGG-16}       & CalTech-101                      &          1             &           1024            &      0.7                                                    &           -            &       -                                                &      -     &         1              &                                                  $2\times2$     &      94.82                                                                          \\ \cline{2-11} 
                              & CalTech-256                      & 1                       &    1024                   &  0.6                                                        &         -              &       -                                                & -           &        1               &      $3\times3$                                                 & 80.16                                                                               \\ \cline{2-11} 
                              & Stanford Dogs                    &     0                  &   -                    & -                                                         &        -               &      -                                                 &          - & -                      &    -                                                   &              80.02                                                                  \\ \hline
\multirow{3}{*}{ResNet-50}    & CalTech-101                      &   1                    &             256          &      0.2                                                    &            2           &   [$3\times3$, $3\times3$]                                                    &  [512, 512]         &     -                  &                        -                               &           92.01                                                                     \\ \cline{2-11} 
                              & CalTech-256                      &      1                 &    1024                   &  0.6                                                        &      1                &      $3\times3$                                                 &     512      &               -        &             -                                          & 82.96                                                                               \\ \cline{2-11} 
                              & Stanford Dogs                    &  1                     & 256                       & 0.4                                                         &        -               &       -                                                &         -  &    -                   &  -                                                     &            81.63                                                                    \\ \hline
\multirow{3}{*}{DenseNet-121} & CalTech-101                      &    1                   &             1024          &      0.3                                                    &               2        &  [$5\times5$, $2\times2$]                                                     & [512, 128]          &    -                   &                       -                                &          95.21                                                                      \\ \cline{2-11} 
                              & CalTech-256                      &      1                 &    1024                   &  0.1                                                        &      2                &      [$2\times2$, $2\times2$]                                                 & [512, 256]  &               -        &             -                                          & 85.44                                                                              \\ \cline{2-11} 
                              & Stanford Dogs                    &  1                     & 512                       & 0.5                                                          &       -                &     -                                                  &        -   & -                      & -                                                      &     83.26                                                                     \\ \hline
\end{tabular}%
}
\end{table*}

\subsubsection{CalTech-256}
The CalTech-256 \cite{griffin2007caltech} dataset is an improvement made over CalTech-101 \cite{fei2004learning} dataset in many aspects such as the total number of images increased from $9144$ to $30607$, the minimum number of images in each class increased from $31$ to $80$, the total number of classes are more than twice the number of images in CalTech-101 dataset, and many more. This dataset has $30, 607$ images, out of which $80\%$ of the images, i.e., $24,485$ are utilized for training the CNNs and remaining images are used to validate the performance of the proposed method. The spatial dimension of the images is re-sized to $224\times224\times3$ to make the image dimension to fit as input to the CNNs. Some sample images of the CalTech-256 dataset are shown in Fig. \ref{fig:dataset_samples}(b).

\subsubsection{Stanford Dogs}
Stanford Dogs dataset \cite{khosla2011novel} consists of $20,580$ images belonging to $120$ different dog breeds. This dataset has developed using the images and annotations from ImageNet \cite{deng2009imagenet} dataset for fine-grained image recognition. To tune and train the CNNs, $12000$ of the total images are utilized and the remaining $8580$ images are used to evaluate the performance of the models. The objects in the images are extracted using the bounding boxes information provided with the dataset. A small set of images from the Stanford Dogs dataset are depicted in Fig. \ref{fig:dataset_samples}(c).



\section{Results and Discussion}
To demonstrate the improved results obtained using the proposed method, this section provides a detailed discussion about the classification performances obtained using the proposed AutoTune method. To find better-performing CNNs, experiments are conducted on three benchmark datasets, including CalTech-101, CalTech-256, and Stanford Dogs. To find the target-specific CNN layers for improved transfer learning, three popular CNNs, including VGG-16 \cite{simonyan2014very}, ResNet-50 \cite{he2016deep}, and DenseNet-121 \cite{huang2017densely} are utilized. It is a standard practice in NAS based works \cite{liu2018progressive, zoph2016neural, white2019bananas} to compare with Random Search (RS). Thus, we also compare the results obtained using the proposed method with the random search approach.
\begin{table*}[!t]
\centering
\caption{The optimal structure of CNN layers found for improved transfer learning using random search. The CNN's layers are tuned w.r.t. the target datasets, such as CalTech-101, CalTech-256, and Stanford Dogs. The optimal configuration of the FC layers does not include the output FC layer.}
\resizebox{\textwidth}{!}{%
\begin{tabular}{|c|c|c|c|c|c|l|l|c|l|c|}
\hline
\multirow{2}{*}{CNN}          & \multirow{2}{*}{Dataset}         & \multicolumn{3}{c|}{FC layer}                                                                            & \multicolumn{3}{c|}{Convolution layer}                                                    & \multicolumn{2}{c|}{Max-pooling layer}                                        & \multirow{2}{*}{\begin{tabular}[c]{@{}c@{}}Validation\\ Accuracy\end{tabular}} \\ \cline{3-10}
                              &                                  & \#layers              & \#neurons             & \begin{tabular}[c]{@{}c@{}}dropout\\ factor\end{tabular} & \#layers              & \begin{tabular}[c]{@{}l@{}}filter\\ size\end{tabular} & \#filters & \#layers              & \begin{tabular}[c]{@{}l@{}}filter\\ size\end{tabular} &                                                                                \\ \hline
\multirow{3}{*}{VGG-16}       & CalTech-101                      &          1             &           512            &      0.6                                                    &           -            &       -                                                &      -     &         -              &                                                  -     &      92.94                                                                          \\ \cline{2-11} 
                              & CalTech-256                      & 1                       &    1024                   &  0.3                                                        &         -              &       -                                                & -           &        1               &      $3\times3$                                                 & 79.99                                                                               \\ \cline{2-11} 
                              & Stanford Dogs                    &     2                  &   [1024, 512]                    & [0.3, 0.5]                                                         &        -               &      -                                                 &          - & -                      &    -                                                   &       77.16                                                                  \\ \hline
\multirow{3}{*}{ResNet-50}    & CalTech-101                      &   1                    &             512          &      0.0                                                    &            2           &   [$2\times2$, $3\times3$]                                                    &  [512, 256]         &     -                  &                        -                               &           90.29                                                                     \\ \cline{2-11} 
                              & CalTech-256                      &      1                 &    1024                   &  0.1                                                        &      -              &      -                                                 &     -      &               -        &             -                                          & 82.84                                                                               \\ \cline{2-11} 
                              & Stanford Dogs                    &  1                     & 1024                       & 0.5                                                         &        -               &       -                                                &         -  &    -                   &  -                                                     &            81.29                                                                    \\ \hline
\multirow{3}{*}{DenseNet-121} & CalTech-101                      &    1                   &             1024          &      0.6                                                    &               1        &  $2\times2$                                                     & 512          &    -                   &                       -                                &          92.55                                                                      \\ \cline{2-11} 
                              & CalTech-256                      &      1                 &    1024                   &  0.2                                                        &      1              &      $3\times3$                                                 & 256  &               -        &             -                                          & 85.16                                                                               \\ \cline{2-11} 
                              & Stanford Dogs                    &  1                     & 512                       & 0.2                                                          &       -                &     -                                                  &        -   & -                      & -                                                      &     82.7                                                                     \\ \hline
\end{tabular}
}
\label{tab:results_random_comparison}
\end{table*}

\subsection{Classification Results over CalTech-101}
The proposed AutoTune method learns the suitable structure of CNN layers for improved transfer learning using  Algorithm \ref{Auto_Tune_algorithm}. Table \ref{tab:results_bayes_comparison} lists the optimal configuration of the hyperparameters involved in CNN layers that are learned using Bayesian Optimization. For instance, fine-tuning the ResNet-50 \cite{he2016deep} (which is pre-trained on ImageNet) over the CalTech-101 dataset with the following structure of CNN layers allows improved transfer learning ability by achieving $92.01\%$ validation accuracy under our experimental settings: 
\begin{itemize}
    \item An additional Fully Connected (FC) layer with  $256$ neurons and having a dropout rate of $0.2$ along with the output FC layer.
    \item The last two convolution layers having $512$ filters of dimension $3\times3$ in each layer. 
\end{itemize}

The proposed AutoTune finds a better configuration of hyperparameters compared to random search. It is evident from the results portrayed in Table \ref{tab:results_random_comparison}. After learning the suitable structure of CNN layers, we fine-tune the learned CNN layers using Adagrad optimizer \cite{duchi2011adaptive} for $200$ epochs. Similar to the setting of training proxy CNNs, we consider the learning rate as $0.01$ and decreased its value with a rate of $\sqrt{0.1}$ for every $50$ epochs. Fine-tuning the base CNN models with the optimal structure of CNN layers which are tuned using Bayesian optimization produce state-of-the-art results while transferring the learned knowledge from source task to the target task. From Table \ref{results_comparison_stateoftheart}, we can observe that the tuning of last few layers of DenseNet \cite{huang2017densely} and VGG-16 \cite{simonyan2014very} using proposed AutoTune method secures the best ($95.92\%$) and the second-best ($95.83\%$) results over the CalTech-101 dataset. 
\begin{table}[]
\caption{The comparison of transfer learning results obtained for CalTech-101 using the conventional fine-tuning and the proposed AutoTune method which is implemented using Bayesian Optimization (BO) and Random Search (RS).}
\label{tab:autotune_caltech101}
\resizebox{0.5\textwidth}{!}{%
\begin{tabular}{|c|c|c|c|c|c|}
\hline
\begin{tabular}[c]{@{}c@{}}CNN \\ Architecture   \end{tabular}           & \begin{tabular}[c]{@{}c@{}}Fine-Tuning\\ Type\end{tabular} & \begin{tabular}[c]{@{}c@{}}\#layers \\ fintuned\end{tabular} & \begin{tabular}[c]{@{}c@{}}\#trainable \\ parameters\\ (in Millions\end{tabular} & \begin{tabular}[c]{@{}c@{}}Validation\\ Accuracy\end{tabular} & \begin{tabular}[c]{@{}c@{}} Total FLOPs (in $10^8$) /\\ FLOPs corresponding \\ to the learned layers\\  (in $10^8$)\end{tabular} \\ \hline
\multirow{3}{*}{VGG-16}       & AutoTune (BO)                                              & 3                                                            & 88.7                                                                             & 95.83                                                         & 153.7/0.51                                                                                                                                                    \\ \cline{2-6} 
                              & AutoTune (RS)                                              & 2                                                            & 26.2                                                                             & 93.54                                                         & 153.5/0.12                                                                                                                                                    \\ \cline{2-6} 
                              & Conventional                                               & 3                                                            & 120                                                                              & 84.21                                                         & 154.6/1.2                                                                                                                                                     \\ \hline
\multirow{3}{*}{ResNet-50}    & AutoTune (BO)                                              & 4                                                            & 4.8                                                                              & 93.57                                                         & 37.6/0.8                                                                                                                                                      \\ \cline{2-6} 
                              & AutoTune (RS)                                              & 4                                                            & 2.4                                                                              & 91.52                                                         & 37.2/0.38                                                                                                                                                     \\ \cline{2-6} 
                              & Conventional                                               & 4                                                            & 4.7                                                                              & 93.55                                                         & 38.5/2.2                                                                                                                                                     \\ \hline
\multirow{3}{*}{DenseNet-121} & AutoTune (BO)                                              & 4                                                            & 1.3                                                                              & 95.92                                                         & 28.4/0.25                                                                                                                                                     \\ \cline{2-6} 
                              & AutoTune (RS)                                              & 3                                                            & 0.6                                                                              & 93.71                                                         & 28.3/0.1                                                                                                                                                      \\ \cline{2-6} 
                              & Conventional                                               & 4                                                            & 0.31                                                                             & 92.94                                                         & 28.2/0.1                                                                                                                                                     \\ \hline
\end{tabular}
}
\end{table}

\begin{table}[]
\caption{Comparing the results obtained for CalTech-256 with the conventional fine-tuning and the proposed AutoTune method which is implemented using Bayesian Optimization (BO) and Random Search (RS).}
\label{tab:autotune_caltech256}
\resizebox{0.5\textwidth}{!}{%
\begin{tabular}{|c|c|c|c|c|c|}
\hline
CNN Architecture              & \begin{tabular}[c]{@{}c@{}}Fine-Tuning\\ Type\end{tabular} & \begin{tabular}[c]{@{}c@{}}\#layers \\ fintuned\end{tabular} & \begin{tabular}[c]{@{}c@{}}\#trainable \\ parameters\\ (in Millions\end{tabular} & \begin{tabular}[c]{@{}c@{}}Validation\\ Accuracy\end{tabular} & \begin{tabular}[c]{@{}c@{}} Total FLOPs (in $10^8$) /\\ FLOPs corresponding \\ to the learned layers\\  (in $10^8$) \end{tabular} \\ \hline
\multirow{3}{*}{VGG-16}       & AutoTune (BO)                                              & 3                                                            & 75.7                                                                             & 82.47                                                         & 153.6/0.2                                                                                                                                                    \\ \cline{2-6} 
                              & AutoTune (RS)                                              & 3                                                            & 42.7                                                                             & 81.83                                                         & 153.6/0.2                                                                                                                                                     \\ \cline{2-6} 
                              & Conventional                                               & 3                                                            & 12.5                                                                             & 79.34                                                         & 154.6/1.2                                                                                                                                                    \\ \hline
\multirow{3}{*}{ResNet-50}    & AutoTune (BO)                                              & 3                                                            & 3.1                                                                              & 84.31                                                         & 38.6/0.6                                                                                                                                                     \\ \cline{2-6} 
                              & AutoTune (RS)                                              & 4                                                            & 2.3                                                                              & 83.74                                                         & 38.5/0.02                                                                                                                                                    \\ \cline{2-6} 
                              & Conventional                                               & 3                                                            & 3.9                                                                              & 83.99                                                         & 38.5/0.03                                                                                                                                                     \\ \hline
\multirow{3}{*}{DenseNet-121} & AutoTune (BO)                                              & 4                                                            & 1.1                                                                              & 86.54                                                        & 28.6/0.5                                                                                                                                                     \\ \cline{2-6} 
                              & AutoTune (RS)                                              & 3                                                            & 0.6                                                                              & 85.96                                                         & 28.3/0.07                                                                                                                                                    \\ \cline{2-6} 
                              & Conventional                                               & 4                                                            & 0.47                                                                             & 84.51                                                         & 28.2/0.1                                                                                                                                                     \\ \hline
\end{tabular}
}
\end{table}

\begin{figure*}[!t]
    \centering
    \includegraphics[width=\textwidth]{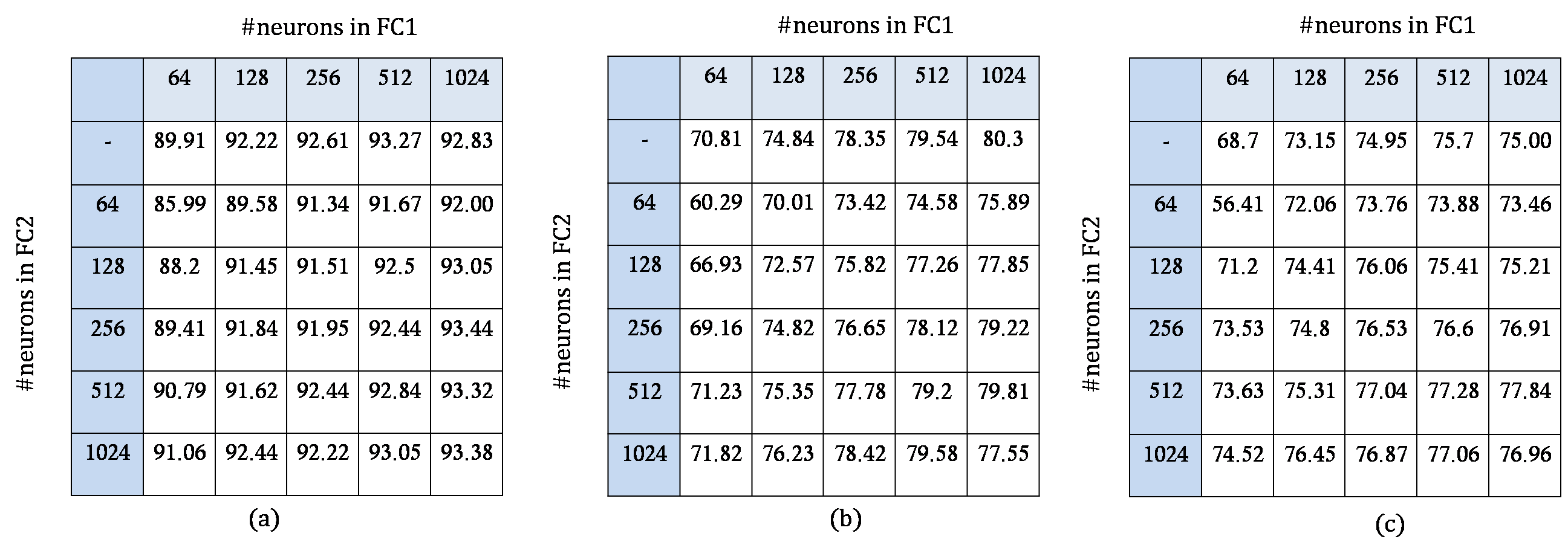}
    \caption{Illustrating the sensitivity of the FC layer's hyperparameters like as the number of FC layers and the number of neurons in FC layers on the performance of the deep neural network. The value of a cell (i, j) represents the validation accuracy obtained by fine-tuning (re-training) the FC layers having the mentioned number of neurons corresponding to row \textit{i} (FC2) and column \textit{j} (FC1). a), b), and c) present the FC layer's configuration of VGG-16 and the validation accuracy obtained over CalTech-101, CalTech-256, and Stanford Dogs datasets, respectively.}
    \label{FC_accuracy}
\end{figure*}

\begin{figure}[!t]
    \centering
    \includegraphics[width=0.5\textwidth]{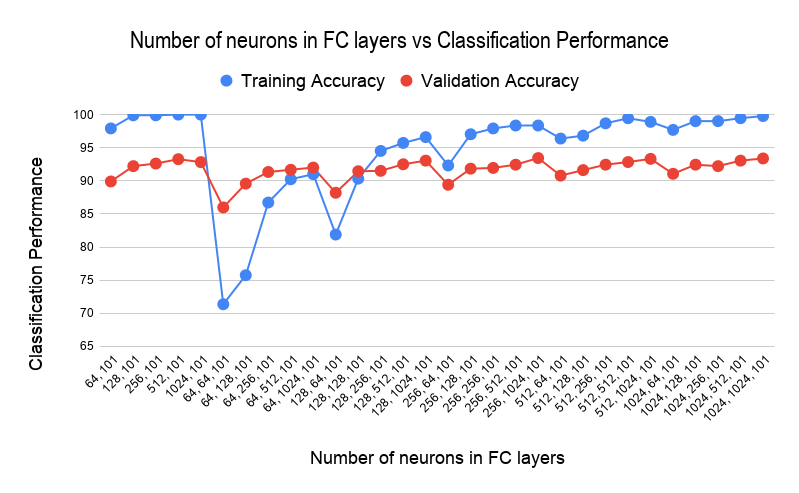}
    \caption{Demonstration of classification performance as a function of the number of neurons in FC layers. The comparison between the training and validation accuracy obtained by re-training the FC layers of VGG-16 with the specified number of neurons over the CalTech-101 dataset.}
    \label{FC_Layers_vs_accuracy}
\end{figure}

\subsection{Classification Results over CalTech-256}
We consider CalTech-256 as another dataset to learn the suitable structure of CNN layers for improved transfer learning. Table \ref{tab:results_bayes_comparison} and \ref{tab:results_random_comparison} presents the optimal structure of CNN layers learned using Bayesian Optimization and random search, respectively. From Table \ref{tab:results_bayes_comparison}, we can note that fine-tuning (training) the CNN with suitable CNN layers over the target dataset enables improved transfer learning ability. For example, fine-tuning the DenseNet-121 \cite{huang2017densely} network with the below structure of CNN layers over the CalTech-256 dataset produces the state-of-the-art results.

\begin{table}[]
\caption{The transfer learning results obtained for Stanford Dogs using the traditional fine-tuning and the proposed AutoTune implemented using Bayesian Optimization (BO) and Random Search (RS).  }
\label{tab:autotune_stanford_dogs}
\resizebox{0.5\textwidth}{!}{
\begin{tabular}{|c|c|c|c|c|c|}
\hline
CNN Architecture              & \begin{tabular}[c]{@{}c@{}}Fine-Tuning\\ Type\end{tabular} & \begin{tabular}[c]{@{}c@{}}\#layers \\ fintuned\end{tabular} & \begin{tabular}[c]{@{}c@{}}\#trainable \\ parameters\\ (in Millions\end{tabular} & \begin{tabular}[c]{@{}c@{}}Validation\\ Accuracy\end{tabular} & \begin{tabular}[c]{@{}c@{}} Total FLOPs (in $10^8$) /\\ FLOPs corresponding \\ to the learned layers\\  (in $10^8$)\end{tabular} \\ \hline
\multirow{3}{*}{VGG-16}       & AutoTune (BO)                                              & 1                                                            & 3                                                                                & 81.23                                                         & 153.4/0.03                                                                                                                                                    \\ \cline{2-6} 
                              & AutoTune (RS)                                              & 3                                                            & 42.7                                                                             & 78.21                                                         & 153.7/0.2                                                                                                                                                     \\ \cline{2-6} 
                              & Conventional                                               & 1                                                            & 0.5                                                                              & 78.25                                                         & 154.6/0.5                                                                                                                                                   \\ \hline
\multirow{3}{*}{ResNet-50}    & AutoTune (BO)                                              & 2                                                            & 0.55                                                                             & 83.17                                                         & 38.5/0.005                                                                                                                                                    \\ \cline{2-6} 
                              & AutoTune (RS)                                              & 2                                                            & 2.2                                                                              & 82.4                                                          & 38.5/0.02                                                                                                                                                     \\ \cline{2-6} 
                              & Conventional                                               & 2                                                            & 1.3                                                                              & 82.21                                                         & 38.5/0.5                                                                                                                                                      \\ \hline
\multirow{3}{*}{DenseNet-121} & AutoTune (BO)                                              & 2                                                            & 0.58                                                                             & 84.67                                                         & 28.2/0.005                                                                                                                                                   \\ \cline{2-6} 
                              & AutoTune (RS)                                              & 2                                                            & 0.5                                                                              & 83.19                                                         & 28.2/0.005                                                                                                                                                    \\ \cline{2-6} 
                              & Conventional                                               & 2                                                            & 0.16                                                                             & 83.35                                                         & 28.2/0.02                                                                                                                                                     \\ \hline
\end{tabular}
}
\end{table}

\begin{itemize}
    \item An additional FC layer with $1024$ neurons and the dropout factor as $0.1$ along with the output FC layer.
    \item The last two convolution layers having filter dimension as $2\times2$ with $512$, $256$ number of filters, respectively.
\end{itemize}

Fine-tuning the proxy CNN with the above specification for DenseNet-121 results in $85.44\%$ classification accuracy, which is state-of-the-art for the CalTech-256 dataset. Furthermore, re-training the DenseNet-121 with the optimal structure of CNN layers for $200$ epochs yields $86.54\%$ classification accuracy. Table \ref{results_comparison_stateoftheart} presents a comparison of the results obtained by employing the proposed AutoTune method with standard baseline results that include both the transfer learning and non-transfer learning based methods. We notice from Table \ref{results_comparison_stateoftheart} that automatically tuning the CNN over the CalTech-256 dataset using DenseNet-121 (Bayesian Optimization)  and DenseNet-121 (Random Search) results in the best ($86.54\%$) and the second-best ($85.96\%$) performance as compared to the state-of-the-art results.

\subsection{Classification Results over Stanford Dogs}
To generalize the significance of the proposed method over different varieties of image classification datasets, we consider the Stanford Dogs dataset having a high degree of inter-class similarity. It is evident from Table \ref{results_comparison_stateoftheart} that the proposed AutoTune method achieves the state-of-the-art result (i.e., $84.67\%$ classification accuracy) over the Stanford Dogs dataset. From this result, we can observe that though the source and target datasets are either from the same domain or from different domains, tuning the CNN w.r.t. the target dataset yields improved transfer learning results using the proposed AutoTune method.

The optimal structure of CNN layers found for Stanford Dogs dataset using Bayesian Optimization and random search methods are shown in Table \ref{tab:results_bayes_comparison} and \ref{tab:results_random_comparison}, respectively. For instance, fine-tuning the ResNet-50 w.r.t. the Stanford Dogs dataset with an additional FC layer having $256$ neurons and dropout factor as $0.4$ along with the output FC layer offers improved performance over validation data. 

\subsection{Comparing the Results with Traditional Transfer Learning}
To demonstrate the improved transfer learning using the proposed method, we compare the results obtained for CalTech-101 using the proposed method with conventional fine-tuning and shown the same in Table \ref{tab:autotune_caltech101}. From this Table, we can observe that the proposed method achieves significant gain in the performance (with a very less number of trainable parameters) compared to the traditional fine-tuning of CNN layers over the target dataset. For instance, traditional fine-tuning of the base VGG-16 model \cite{simonyan2014very} achieves $84.21\%$ validation accuracy for CalTech-101, which requires to train $120$ Million parameters involved in top $3$ layers. On the other hand, fine-tuning the VGG-16 model tuned using the proposed AutoTune method attains $95.83\%$ validation accuracy by training $88.7$ parameters, which is $0.26$ times fewer than the parameters involved in traditional fine-tuning. 

We have also reported the fine-tuning results obtained for CalTech-256 and Stanford Dogs datasets in Table \ref{tab:autotune_caltech256} and Table \ref{tab:autotune_stanford_dogs}, respectively. We can observe from these tables that the trainable parameters and FLOPs resulted with the proposed method are less than the conventional method in most of the cases. It happens because the last few layers are learned w.r.t. the target dataset that has less number of training images. On the other hand, in a few cases (i.e., ResNet-50 and DenseNet-121), the resulted trainable parameters with our method are slightly high since we consider additional FC layers in the search space. Furthermore, to demonstrate the sensitivity of the hyperparameters such as the number of neurons in FC layers, we conduct experiments with VGG-16 by varying the number of FC layers, the number of neurons in each FC layer. The results are illustrated in Fig.\ref{FC_accuracy}. From this figure, we can observe the validation accuracies obtained for CalTech-101, CalTech-256, and Stanford Dogs datasets using different configurations of FC layers. For example, fine-tuning (re-training) the weights involved in FC1, FC2, and output FC layers with 512, 1024, and 101 neurons over the CalTech-101 dataset results in $93.05\%$ classification performance. Note that the dropout factor is considered as 0.5 after every FC or Dense layer except the output FC layer. The significance of learning the CNN architecture with the knowledge of target dataset is illustrated in Fig. \ref{FC_Layers_vs_accuracy}. The sensitivity of the FC layer's hyperparameters can be observed over the classification performance in Fig. \ref{FC_Layers_vs_accuracy}. The above analysis shows that choosing the suitable FC layer's configuration leads to improve the generalization ability of the deep neural network over the target dataset. 

\section{Conclusion and Future Work}
We propose a novel framework for automatically tuning the pre-trained CNN w.r.t. the target dataset while transferring the learned knowledge from the source task to the target task. We compare the Bayesian and Random search strategies to perform the tuning of network hyperparameters. Experiments are conducted using VGG-16, ResNet-50, and DenseNet-121 models over the CalTech-101, CalTech-256, and Stanford Dogs datasets. The models are originally trained over the large-scale ImageNet dataset. The experimental results suggest that the tuning of the CNN layers with the knowledge of the target dataset improves the transferring ability. Automatically tuning the CNN by utilizing the knowledge from the target dataset demonstrates a significant reduction in the error by $27.4\%$, $17.9\%$, and $5.6\% $ over the CalTech-101, CalTech-256, and Stanford Dogs datasets, respectively. We further observe that the proposed AutoTune improves performance using both Bayesian and Random search strategies. Moreover, the proposed AutoTune performs much better than the conventional fine-tuning of transfer learning as depicted by the results reported in this study. The FLOPs corresponding to the CNN that is learned using the proposed approach are also optimum. In the future, we would like to extend this idea by employing filter pruning methods after finding the suitable CNN architecture to discover a light-weight CNN for efficient transfer learning.

\section*{Acknowledgments}
We are grateful to NVIDIA corporation for supporting us by donating an NVIDIA Titan Xp GPU, which is used for this research.

\bibliographystyle{IEEEtran} 
\bibliography{refs}
\end{document}